\documentclass[a4paper]{article}

\usepackage[utf8]{inputenc}
\usepackage{erk}
\usepackage{times}
\usepackage{graphicx}
\usepackage[top=22.5mm, bottom=22.5mm, left=22.5mm, right=22.5mm]{geometry}

\usepackage[pagebackref=false,breaklinks=true,colorlinks,bookmarks=false]{hyperref}

\usepackage[slovene,english]{babel}

\usepackage{amsmath}
\usepackage{bm}
\usepackage{indentfirst}
\usepackage{multirow}
\usepackage{multicol}
\usepackage[table]{xcolor}
\usepackage{hhline}

\def\footnotemark{}

\begin{document}
\title{
Iterativna optimizacija ocen kakovosti slikovnih podatkov v sistemih za razpoznavanje obrazov
}

\author{Žiga Babnik, Vitomir Štruc} 

\thanks{
Podprto s strani ARRS raziskovalnega programa P2--0250 (B), ter ARRS programom mladih raziskovalcev.
}

\affiliation{Univerza v Ljubljani, Fakulteta za Elektrotehniko,
Tržaška cesta 25, 1000 Ljubljana, Slovenija
}

\email{\{ziga.babnik, vitomir.struc\}@fe.uni-lj.si}

\maketitle

\begin{abstract}{Iterative Optimization of Pseudo Ground-Truth Face Image Quality Labels\vspace{3mm}}
While recent face recognition (FR) systems achieve excellent results in many deployment scenarios, their performance in challenging real-world settings is still under question. For this reason, face image quality assessment (FIQA) techniques aim to support FR systems, by providing them with sample quality information that can be used to reject poor quality data unsuitable for recognition purposes. Several groups of FIQA methods relying on different concepts have been proposed in the literature, all of which can be used for generating quality scores of facial images that can serve as pseudo ground-truth (quality) labels and can be exploited for training (regression-based) quality estimation models.   
Several FIQA appro\-aches show that a significant amount of sample-quality information can be extracted from mated similarity-score distributions generated with some face matcher. Based on this insight, we propose in this paper a quality label optimization approach, which incorporates sample-quality information from mated-pair similarities into quality predictions of existing off-the-shelf FIQA techniques. We evaluate the proposed approach using three state-of-the-art FIQA methods over three diverse datasets. The results of our experiments show that the proposed optimization procedure heavily depends on the number of executed optimization iterations. At ten iterations, the approach seems to perform the best, consistently outperforming the base quality scores of the three FIQA methods, chosen for the experiments.
\end{abstract}

\selectlanguage{slovene}

\section{Uvod}\label{sec:introduction}

Moderni sistemi za razpoznavanje obrazov dosegajo izvrstne rezultate, tudi na večjih, ter bolj težavnih podatkovnih zbirkah obraznih slik, kot je recimo zbirka IARPA Janus Bench\-mark-C (IJB-C)~\cite{ijb-c}. Vendar je prenos izjemnih rezultatov v realni svet, za naloge kot je video nadzor, zaradi slabe kakovosti slikovnih podatkov zaenkrat še neuresničljiv. Za lažje uresničevanje doseganja dobrih in zanesljivih rezultatov obraznih razpoznavalnikov, se je pojavilo raziskovalno področje ocenjevanja kakovosti obraznih slik (angl. Face Image Quality Assessment - FIQA), ki skuša oceniti biometrično kakovost vzorca za namene razpoznavanja~\cite{isoiec}. Biometrična kakovost je pogosto definirana kot \textit{koristnost oz. primernost} vzorca za namene obdelave v sodobnih sistemih za razpoznavanje obrazov ~\cite{surveypaper}. Biometrična kakovost je torej tesno povezana z vizualno kakovostjo vzorca, a ji ni povsem enaka.

Obstaja več skupin pristopov ocenjevanja (biometr\-ične) kakovosti obraznih slik. Najbolj razširjena skupina se poslužuje ustvarjanja psevdo referenčnih vrednosti kakovosti večjega nabora vzorcev. Referenčne vrednosti nato uporabijo za učenje regresijskih modelov, sposobnih samostojnega napovedovanja ocen kakovosti~\cite{faceqnet, sddfiqa, pcnet, lightqnet}. V zadnjem času se pojavljajo tudi metode, ki združujejo nalogo razpoznavanja obrazov in ocenjevanja kakovosti~\cite{magface, pfe}, ter analitični postopki, ki za oceno kakovosti uporabijo zgolj karakteristike vhodnega vzorca in lastnosti izbranega razpoznavalnika~\cite{serfiq, faceqan}.

Pristope iz katerekoli zgoraj omenjene skupine je mo\-goče uporabiti tudi za razvoj FIQA metod z nadzorovanim učenjem. V tem primeru za večji nabor obraznih slik s pomočjo izbrane FIQA metode pridobimo ocene (psevdo) referenčnih oznak kakovosti, le-te pa nato uporabimo za učenje modela za ocenjevanje kakovosti. Pri tem lahko pridobljene ocene kakovosti pred učenjem obogatimo in izboljšamo s pomočjo dodatnih zunanjih informacij, kot je recimo porazdelitev podobnosti med ujemajočimi se pari obraznih slik (v smislu identitete). Mno\-gi obstoječi FIQA pristopi namreč kažejo na dejstvo, da podobnost takšnih parov vsebuje precejšnjo količino informacij o sami kakovosti posameznih vzorcev~\cite{sddfiqa, lightqnet}. Na podlagi predstavljene ideje v tem članku predstavimo pristop za (iterativno) izboljševanje začetnih ocen kakovosti obraznih slik, ustvarjenih z izbranim FIQA modelom, ki temelji na vključevanju dodatnih informacij, pridobljenih iz primerjav ujemajočih se slik obrazov. Izboljšane ocene lahko nato uporabimo v postopku nadzorovanega učenja regresijske mreže za ocenjevanje kakovosti obraznih slik.

\section{Pregled področja}\label{sec:related_work}

V tem razdelku predstavimo tri glavne skupine FIQA metod, ki jih lahko razdelimo v: 
analitične, regresijske ter mrežne pristope. Podrobnejši pregled področja je predstavljen v nedavnem preglednem članku~\cite{surveypaper}.

\textbf{Analitični pristopi} temeljijo na izločanju ocene kakovosti iz informacije prisotne v samem vzorcu. Zaradi tega se pretežno osredotočajo na vizualno kakovost in pogostjo ne dosegajo konkurenčnih rezultatov v primerjavi z najnaprednejšimi rešitvami iz literature. Starješi pristop, ki so ga predstavili Gao \textit{et al.}~\cite{anal1} poskuša pridobiti oceno kakovosti slikovnih podatkov z ocenjevanjem obrazne simetrije. Pred kratkim sta se pojavila dva pristopa, ki poleg karakteristik slik hkrati upoštevata tudi informacijo izbranega razvrščevalnika, in tako dosegata vrhunske rezultate. Prvi pristop, ki so ga predstavili Terh\-örst \textit{et al.}~\cite{serfiq}, se zanaša na uporabo izpustnih slojev mrež, medtem ko drugi, predlagan s strani Babnika \textit{et al.}~\cite{faceqan}, za oceno kakovosti izrablja nasprotniške pristope.

\textbf{Regresijski pristopi} predstavljajo najštevilčnejšo sk\-upino FIQA metod, ki temeljijo na učenju regresijksih modelov za ocenjevanje kakovosti iz pridelanih psevdo referenčnih oznak kakovosti vzorcev. Eden izmed zače\-tnih postopkov iz te skupine, Ortege \textit{et al.}~\cite{faceqnet}, uči regresijsko nevronsko mrežo, z uporabo oznak, pridobljenih s primerjavo vložitev najkakovostnejši sliki vsakega posameznika. Pri tem se najkakovostnejše slike posameznikov izloči s pomočjo zunanjega orodja za preverjanje kakovosti. Naprednješi pristop, ki so ga predlagali Ou \textit{et al.}~\cite{sddfiqa}, za napoved referenčnih oznak kakovosti uporabi podobnosti tako ujemajočih kot tudi neujemajočih se parov obraznih slik  in s tem pridela koristnejše ocene kakovosti za učenje regresijskega FIQA modela.

\textbf{Mrežni pristopi} največkrat združijo nalogi razpoznavanja ter ocenjevanja kakovosti in se naučijo preslikave vhodnih slik v posebne vložitve, iz katerih je možno izlo\-čiti informacijo o identiteti, kot tudi informacijo o kakovosti vzorca. Starejši pristop avtorjev Shi in Jain~\cite{pfe} se nauči napovedati dvojice vektorjev, kjer prvi -- povprečni vektor vsebuje informacijo o identiteti, drugi -- vektor odklona, iz katerega je možno izraziti kakovost vzorca, pa se navezuje na nedoločenost povprečnega vektorja. Novejši pristop, predlagan s strani Meng \textit{et al.}~\cite{magface}, za potrebe učenja razpoznavalnika obrazov uporabi prirejeno ArcFace funkcijo izgube, ki vključuje tudi informacijo o kakovosti vhodnega vzorca. 

\section{Metodologija}\label{sec:methodology}

Poglavitna naloga metod za oceno kakovosti obraznih slik je zajetje čim večje količine informacij o koristnosti oz. primernosti vzorca za biometrično razpoznavanje v končno oceno kakovosti. Mnoge raziskave kažejo na dejstvo, da podobnosti ujemajočih se parov obraznih slik vsebujejo velike količine informacij o kakovosti vzorca, zato v tem razdelku predstavimo pristop, ki na podlagi teh informacij, iterativno optimizira predhodne napovedi kakovosti slikovnih vzorcev. Izboljšane ocene kakovosti lahko nato uporabimo v postopku učenja regresijske mreže za ocenjevanje kakovosti obraznih slik.

\subsection{Pregled pristopa}\label{sec:introduction:overview}

Predpostavimo poljuben FIQA pristop za oceno kakovosti obraznih slik $F_q$, ki za podani obrazni vzorec $I_i$ vrne oceno kakovosti $q_{I_i} = F_q(I_i)$, večjo podatkovno zbirko obraznih slik $\mathcal{I}=\{I_i\}_{i=1}^n$, ki vsebuje slike $N\leq n$ različnih posameznikov, ter ciljni obrazni razpoznavalnik $M$. Cilj našega pristopa je pridobiti optimizirane ocene kakovosti $\mathcal{Q}^*=\{q_{I_i}^*\}_{i=1}^n$, z iterativnim posodabljanjem osnovnih ocen kakovosti $\mathcal{Q}=\{q_{I_i}\}_{i=1}^n$, ki nam jih vrne pristop za oceno kakovosti $F_q$, na podlagi informacij pridobljenih iz porazdelitve podobnosti ujemajočih se slik obrazov $\mathcal{D}=\{d_i\}_{i=1}^m$, kjer je $m$ število upoštevanih primerjav slik.

\subsection{Inicializacija pristopa}\label{sec:introduction:initialization}

Preden lahko izvedemo optimizacijo ocen, jih je potrebno najprej pridobiti. Zato uporabimo izbran pristop $F_q$ in ga izvedemo nad celotno zbirko podatkov $\mathcal{I}$. Tako pridobimo osnovne ocene $\mathcal{Q} = F_q(\mathcal{I})$, ki jih nato normiramo na interval $[0, 1]$. Prav tako potrebujemo vnaprej ustvarjene vložitve celotne zbirke $\mathcal{I}$, tj. $\mathcal{E} = M(\mathcal{I})$, ki jih pridobimo s pomočjo razpoznavalnika obrazov $M$ . 

Na podlagi izbranega nabora $\mathcal{I}$ in pripadajočega seznama identitet $\{C_i\}_{i=1}^N$, zgradimo seznam ujemajočih se slik, tako da za vsak vzorec iz zbirke izberemo $k$ naključnih vzorcev iste identitete. Za posamezen pristni par vzorcev $(I_i, I_j)$, kjer velja $I_i,I_j\in C_k$, nato izračunamo podobnost z uporabo kosinusne podobnosti:
\begin{equation}\label{eq:cosine_similarity}
    s_{cos}(I_i, I_j) = \frac{e_i \cdot e_j}{\|e_i\| \cdot \|e_j\|},
\end{equation}
kjer sta $e_i = M(I_i)$ ter $e_j = M(I_j)$ vložitvi prej omenjenih vzorcev. Izračunane podobnosti ležijo na intervalu $[-1, 1]$, zato jih  podobno kot ocene kakovosti normiramo na interval $[0, 1]$. Z izračunom podobnosti med vsemi pristnimi dvojicami lahko ustvarimo porazdelitev podobnosti pristnih dvojic $\mathcal{D}$. Za lažje razumevanje nadaljnjih korakov je potrebno izpostaviti še definicijo kakovosti para slik $(I_{I_i}, I_{I_j})$, ki je pogosto uporabljena na področju ocenjevanja kakovosti:
\begin{equation}\label{eq:pair_similarity}
    q_p(I_i, I_j) = min(q_{I_i}, q_{I_j}),
\end{equation}
ki pravi, da je kakovost para kar enaka kakovosti slabše slike v paru.

\subsection{Iterativna optimizacija}\label{sec:introduction:iteration}

Po inicializaciji podobnosti pristnih dvojic $\mathcal{D}$, torej parov slik, ki pripadajo isti osebi, lahko izvedemo posamezno iteracijo predlaganega pristopa. Prvi korak vključuje izračun korekcijskega faktorja $\theta_{I_i}$ vzorca $I_i$:
\begin{equation}\label{eq:correction_factor}
    \theta_{I_i} = \frac{1}{k} \sum_{j=0}^{k} s_{cos}(I_i, I_j),
\end{equation}
kjer je $(I_i, I_j)$ pristni par . Pri izračunu uporabimo samo pare za katere velja
\begin{equation}
    \quad q_{I_i} \leq q_{I_j} - \lambda,
\end{equation}
da je ocena kakovosti prve slike $q_{I_i}$ manjša za več kot $\lambda$ od ocene kakovosti druge slike $q_{I_j}$.
Zaradi prej omenjene definicije kakovosti para vzorcev, prikazane v Enačbi~\eqref{eq:pair_similarity}, ki pravi, da na kakovost para vzorcev vpliva le manj kakovostni vzorec v paru, uporabimo pri izračunu korekcijskega faktorja vzorca $I_i$ le pare, kjer $I_i$ nastopa kot manj kakovosten vzorec. Korekcijski faktor $\theta_{I_i}$ predstavlja oceno kakovosti vzorca $I_i$, ki upošteva le informacijo prisotno iz porazdelitve pristnih parov.
S pomočjo izračunanega korekcijskega faktorja lahko posodobimo trenutno kakovost vzorca posamezne slike $I_i$ kot
\begin{equation}\label{eq:update_function}
    q_{I_i}^{j+1} = q_{I_i}^{j} + \epsilon \cdot (\theta_{I_i} - q_{I_i}^{j}),
\end{equation}
kjer je $j$ trenutna iteracija, ter $\epsilon$ vhodni parameter pristopa, ki določa velikost premika posamezne iteracije. S premikom osnovne ocene, proti vrednosti korekcijskega faktorja v oceno vključimo dodatno informacijo pridobljeno iz porazdelitve pristnih parov.
Po izvedbi $L$ iteracij, pridobimo nove, optimizirane ocene kakovosti $\mathcal{Q}^*=\{q_{I_i}^*\}_{i=1}^n$, kjer je $q_{I_i}^* = q_{I_i}^{L}$. 

Ker lahko z naključnim izbiranjem pristnih dvojic v pristop vnesemo pristranskost, celoten postopek ponov\-imo v $T$ ponovitvah. Tako končno oceno kakovosti izraču\-namo kot
\begin{equation}\label{eq:final_score}
    \mathcal{Q}_{I_i} = \frac{1}{T} \sum_{t=0}^{T} {\mathcal{Q}_{I_i}^t}^*,
\end{equation}
kjer je ${\mathcal{Q}_{I_i}^t}^*$ množica optimiziranih ocen kakovosti $t$-te ponovitve postopka.

\section{Eksperimenti in rezultati}\label{sec:experiments_results}

\subsection{Zasnova eksperimentov}\label{sec:experiments_results:experiments}

\renewcommand{\footnotesize}{\scriptsize}

\begin{table*}[!th]
    \centering
    
    \caption{Primerjava vrednosti površine pod krivuljo ($\downarrow$) za vse izbrane metode na treh izbranih podatkovnih zbirkah. Najslabši rezultat znotraj posamezne podatkovne zbirke, metode in deleža zavrnjenih slik je obarvan rdeče, najboljši zeleno.\vspace{1mm}}
    \label{tab:my_label}
    
    \resizebox{\linewidth}{!}{%
    \begin{tabular}{|c|c  |r|r|r|r||r|  |r|r|r|r||r|  |r|r|r|r||r|}

\hline
 \multicolumn{2}{|c|}{} & \multicolumn{5}{c||}{\textbf{CPLFW}} & \multicolumn{5}{c||}{\textbf{CALFW}} & \multicolumn{5}{c|}{\textbf{XQLFW}}\\ 
\hline
\textbf{M}$^\dagger$ & \textbf{I}$^\ddagger$ & \bm{$10\%$} & \bm{$20\%$} & \bm{$40\%$} & \bm{$80\%$} & \bm{$\mu$} & \bm{$10\%$} & \bm{$20\%$} & \bm{$40\%$} & \bm{$80\%$} & \bm{$\mu$} & \bm{$10\%$} & \bm{$20\%$} & \bm{$40\%$} & \bm{$80\%$} & \bm{$\mu$}\\ 

\hline
\parbox[t]{2mm}{\multirow{4}{*}{\rotatebox[origin=c]{90}{\textbf{FaceQAN}}}}
 & 0 & $1.417$ & $2.226$ & $3.511$ & $5.155$ & $3.077$
        & {\cellcolor{green!10}}$1.512$ & $2.737$ & $4.594$ & {\cellcolor{green!10}}$7.939$ & $4.195$
        & $5.148$ & $8.882$ & {\cellcolor{red!10}}$13.2$ & {\cellcolor{red!10}}$16.325$ & {\cellcolor{red!10}}$10.888$ \\
 & 5 & {\cellcolor{red!10}}$1.424$ & {\cellcolor{red!10}}$2.229$ & {\cellcolor{red!10}}$3.555$ & {\cellcolor{red!10}}$5.303$ & {\cellcolor{red!10}}$3.127$
         & $1.524$ & $2.782$ & {\cellcolor{red!10}}$4.845$ & {\cellcolor{red!10}}$8.139$ & {\cellcolor{red!10}}$4.322$
         & $5.14$ & {\cellcolor{green!10}} $8.56$ & {\cellcolor{green!10}} $12.998$ & {\cellcolor{green!10}} $16.046$ & {\cellcolor{green!10}}\bm{$10.686$} \\
 & 10 & {\cellcolor{green!10}} $1.407$ & $2.203$ & $3.492$ & {\cellcolor{green!10}} $4.788$ & {\cellcolor{green!10}}\bm{$2.972$}
         & $1.523$ & {\cellcolor{green!10}} $2.72$ & {\cellcolor{green!10}} $4.587$ & $7.941$ & {\cellcolor{green!10}}\bm{$4.192$}
         & {\cellcolor{green!10}} $5.076$ & $8.75$ & $13.083$ & $16.246$ & $10.788$\\
 & 15 & $1.414$ & {\cellcolor{green!10}} $2.201$ & {\cellcolor{green!10}} $3.489$ & $4.891$ & $2.998$
         & {\cellcolor{red!10}}$1.532$ & {\cellcolor{red!10}}$2.847$ & $4.69$ & $8.113$  & $4.295$
         & {\cellcolor{red!10}}$5.169$ & {\cellcolor{red!10}}$8.883$ & $13.143$ & $16.199$ & $10.848$ \\
\hhline{*{17}{-}}
\parbox[t]{2mm}{\multirow{4}{*}{\rotatebox[origin=c]{90}{\textbf{CR-FIQA}}}}
 & 0 & $1.408$ & $2.193$ & $3.131$ & $4.373$ & $2.776$
        & $1.485$ & {\cellcolor{red!10}}$2.773$ & $5.034$ & $8.327$ & $4.404$
        & $5.136$ & $9.171$ & {\cellcolor{red!10}}$14.874$ & {\cellcolor{red!10}}$18.333$ & {\cellcolor{red!10}}$11.878$ \\
 & 5 & {\cellcolor{red!10}}$1.41$ & {\cellcolor{red!10}}$2.197$ & $3.133$ & $4.408$ & $2.787$
        & {\cellcolor{red!10}}$1.486$ & $2.77$ & {\cellcolor{red!10}}$5.052$ & {\cellcolor{red!10}}$8.328$ & {\cellcolor{red!10}}$4.409$
        & {\cellcolor{green!10}} $5.135$ & {\cellcolor{green!10}} $9.169$ & {\cellcolor{green!10}} $14.074$ & $17.543$ & {\cellcolor{green!10}}\bm{$11.480$} \\
 & 10 & {\cellcolor{green!10}} $1.404$ & {\cellcolor{green!10}} $2.191$ & {\cellcolor{red!10}}$3.327$ & {\cellcolor{red!10}}$4.577$ & {\cellcolor{red!10}}$2.874$
        & {\cellcolor{green!10}} $1.473$ & $2.754$ & $4.916$ & {\cellcolor{green!10}} $8.008$ & {\cellcolor{green!10}}\bm{$4.287$}
        & $5.266$ & $9.297$ & $14.184$ & {\cellcolor{green!10}} $17.435$ & $11.545$\\
 & 15 & $1.408$ & $2.192$ & {\cellcolor{green!10}} $3.105$ & {\cellcolor{green!10}} $4.334$ & {\cellcolor{green!10}}\bm{$2.759$}
        & $1.477$ & {\cellcolor{green!10}} $2.74$ & {\cellcolor{green!10}} $4.873$ & $8.086$  & $4.294$
        & {\cellcolor{red!10}}$5.277$ & {\cellcolor{red!10}}$9.436$ & $14.456$ & $17.928$ & $11.774$\\
\hhline{*{17}{-}}
\parbox[t]{2mm}{\multirow{4}{*}{\rotatebox[origin=c]{90}{\textbf{SDD-FIQA}}}}
 & 0 & {\cellcolor{red!10}}$1.517$ & {\cellcolor{red!10}}$2.467$ & $3.47$ & $4.615$ & $3.017$
        & $1.465$ & {\cellcolor{red!10}}$2.719$ & {\cellcolor{red!10}}$4.922$ & $8.248$ & {\cellcolor{red!10}}$4.338$
        & $5.214$ & $9.414$ & {\cellcolor{red!10}}$16.695$ & {\cellcolor{red!10}}$22.968$ & {\cellcolor{red!10}}$13.572$ \\
 & 5 & $1.505$ & $2.431$ & {\cellcolor{green!10}} $3.404$ & {\cellcolor{green!10}} $4.586$  & {\cellcolor{green!10}}\bm{$2.981$}
        & {\cellcolor{green!10}} $1.45$ & $2.705$ & $4.803$ & {\cellcolor{red!10}}$8.375$ & $4.333$
        & {\cellcolor{green!10}} $5.165$ & {\cellcolor{green!10}} $9.406$ & $16.19$ & $21.711$ & {\cellcolor{green!10}}\bm{$13.118$}\\
 & 10 & {\cellcolor{green!10}} $1.482$ & {\cellcolor{green!10}} $2.372$ & $3.544$ & $4.671$ & $3.017$
        & {\cellcolor{red!10}}$1.466$ & $2.715$ & $4.874$ & $8.122$ & $4.294$ 
        & $5.43$ & $9.756$ & {\cellcolor{green!10}} $16.07$ & {\cellcolor{green!10}} $21.498$ & $13.188$ \\
 & 15 & $1.483$ & $2.382$ & {\cellcolor{red!10}}$3.587$ & {\cellcolor{red!10}}$4.756$  & {\cellcolor{red!10}}$3.052$
        & $1.461$ & {\cellcolor{green!10}} $2.705$ & {\cellcolor{green!10}} $4.725$ & {\cellcolor{green!10}} $8.08$ & {\cellcolor{green!10}}\bm{$4.242$}
        & {\cellcolor{red!10}}$5.439$ & {\cellcolor{red!10}}$9.813$ & $15.965$ & $21.748$ & $13.241$ \\
\hhline{*{17}{-}}

\multicolumn{8}{l}{\textbf{M}$^\dagger$ - FIQA metoda , \textbf{I}$^\ddagger$ - število iteracij} \\
\end{tabular}
}
\begin{tabular}{c c c c}
     \cellcolor{green!10} & - najboljši rezultat, & {\cellcolor{red!10}} & - najslabši rezultat \\
\end{tabular}
\end{table*}

\textbf{Modeli in podatki.} Za začetne ocene kakovosti obraznih slik smo uporabili tri napredne FIQA pristope : {CR-FIQA}\footnote{\label{note1}\url{https://github.com/fdbtrs/CR-FIQA}}~\cite{crfiqa}, {FaceQAN}\footnote{\label{note2}\url{https://github.com/LSIbabnikz/FaceQAN}}~\cite{faceqan} ter {SDD-FIQA}\footnote{\label{note3}\url{https://github.com/Tencent/TFace/tree/quality}}~\cite{sddfiqa}. Za potrebe učenja in testiranja smo izbrali obrazni razpoznavalnik {ArcFace}\footnote{\label{note3}\url{https://github.com/deepinsight/insightface}}~\cite{arcface}, ki temelji na ResNet100 arhitekturi, učeni z izgubo kotnega razmaka. Za učenje smo izbrali zbirko {VGGFace2}~\cite{vggface2}, ki vsebuje približno $3$ milijone slik, $9131$ različnih identitet, za testiranje pa zbirke: {CPLFW}~\cite{cplfw}, {CALFW}~\cite{calfw} ter {XQLFW}~\cite{xqlfw}. Podatkovna zbirka CPLFW se osredotoča na razlike v obrazni pozi, CALFW na razlike v starosti, XQLFW pa vsebuje obraze s širokim razponom v (vizualni) kakovosti slik.

\textbf{Ovrednotenje zmogljivosti.} Omenjeni razpoznavalnik ArcFace smo priredili za potrebe ocenjevanja kakovosti tako, da smo dodali regresijsko glavo in ga učili s pomočjo $L_1$ funkcije izgube. Učenje smo izvedli dvakrat, prvič na osnovnih/začetnih ocenah kakovosti in drugič na optimiziranih ocenah kakovosti vseh pristopov. Za potrebe primerjave zmogljivosti mrež učenih iz osnovnih in optimiziranih ocen smo uporabili ustaljen pristop krivulj zmote zoper zavrnitve, ki merijo stopnjo lažno negativnih parov pri vnaprej določeni vrednosti stopnje lažno pozitivnih parov kot funkcijo različnih stopenj zavrnitve slik zaradi nizke kakovosti \cite{faceqan}. Za pridobljene krivulje izračunamo ploščino pod krivuljo, kjer manjša vrednost namiguje na boljše ocene kakovosti. Za namene ovrednotenja optimizacijskega postopka, v članku prikažemo ploščino pod krivuljo pri različnih stopnjah zavrnitve.

\textbf{Podrobnosti o implementaciji.} Za zagotavljanje robustnih ocen kakovosti, smo s pomočjo preliminarnih eksperimentov določili $T=10$ za število ponovitev, pri čemer za vsako sliko v posamezni ponovitvi zgradimo $N=10$ pristnih parov. Hkrati tako zagotovimo robustnost ter omejimo časovno kompleksnost pristopa. Cilj samega pristopa je le natančnejša prilagoditev začetnih ocen kakovosti, zato izberemo $\epsilon=0.01$ in tako močno omejimo korak posamezne iteracije metode. Določimo še $\lambda=0.05$ in tako dodatno kaznujemo pare slik, kjer sta kvaliteti pod zastavljeno mejo. Za število iteracij $L$ izberemo tri različne vrednosti $L \in [5, 10, 15]$, saj nas zani\-ma učinek povečevanja števila iteracij na končne rezultate. Eksperimenti so izvedeni na namiznem računalniku z Intel i9-10900KF procesorjem, 64GB procesorskega pomnilnika ter z Nvidia 3090 grafično kartico, za izračun posamezne iteracije na podatkovni bazi VGGFace2, iz katere sestavimo okoli 33 milijonov pristnih parov, potrebujemo približno 540 sekund.

\subsection{Primerjava dobljenih rezultatov}\label{sec:experiments_results:results}

Tabela~\ref{tab:my_label} prikazuje rezultate eksperimentov, natančn\-eje, vrednosti površine pod krivuljo za vse tri izbrane pristope ocenjevanja kakovosti obraznih slik. Za realne aplikacije so bolj pomembni rezultati pri manjših deležih zavrnjenih slik, saj  zaradi težav s kakovostjo ne želimo zavrniti prevelikega deleža slik. V tabeli~\ref{tab:my_label} so zato prikazane vrednosti površin pri $10\%$, $20\%$, $40\%$ ter $80\%$ zavrnjenih slik. Posamezne vrstice rezultatov so označene s številom izvedenih iteracij, $0$ iteracij predstavlja osnovne rezultate pristopa brez optimizacije. Za lažjo predstavitev in interpretacijo rezultatov predstavimo tudi povprečno vrednost za vse stopnje izpuščenih slik -- označeno z $\mu$.

\textbf{FaceQAN.} Opazimo lahko, da FaceQAN najboljše rezultate v povprečju doseže pri desetih iteracijah, kjer pridobimo najboljše rezultate za zbirkah CPLFW ter CA\-LFW. Na zbirki XQLFW pa boljše rezultate pristop dose\-že le še pri petih iteracijah. Kljub rezultatom na bazi XQLFW, metoda v povprečju preko vseh podatkovnih zbirk pridela najslabše rezultate prav pri petih iteracijah, ki so slabši tudi od osnovnih ocen kakovosti metode.

\textbf{CR-FIQA.} Rezultati pristopa CR-FIQA preko različ\-nih podatkovnih zbirk kažejo zelo raznolike rezultate, ki so močno odvisni od karakteristik testnih podatkov. Za bazo CPLFW dosegajo najboljše rezultate ocene pri petnajstih iteracijah, za CALFW pri desetih in za XQLFW pri petih iteracijah. Z izjemo desetih iteracij na bazi CP\-LFW, dosegajo ocene desetih in petnajstih iteracij dosledno boljše rezultate kot osnovne ocene. Kljub izjemi na bazi CPLFW lahko opazimo, da najboljše rezultate v povprečju preko vseh zbirk dosegajo ocene desetih iteracij.

\textbf{SDD-FIQA.} Rezultati pristopa SDD-FIQA kažejo, da predstavljena metoda optimizacije uspe izboljšati ocene kakovosti v vseh primerih razen na podatkovni bazi CPL\-FW v primeru petnajstih iteracij, kjer je rezultat slabši od osnovnih ocen pristopa. Kjub temu, da v primeru desetih iteracij ne dosežemo najboljšega rezultata za nobeno izmed podatkovnih zbirk, so rezultati v tem primeru najboljši. Tesno sledijo rezultati petih in petnajstih iteracij, najslabše rezultate pa dosegajo osnovne ocene metode.

\textbf{Primerjava med pristopi.} Iz rezultatov je razvidno, da lahko z izbiro pravega števila iteracij pridobimo boljše ocene kakovosti na posamezni podatkovni zbirki, kot jih zagotavljajo osnovni FIQA pristopi. Izboljšava je sicer majhna, kar ni nepričakovano, saj je namen pristopa le optimizacija predhodnih ocen kakovosti vzorcev, a je lah\-ko pomembna za realne sisteme razpoznavanja obrazov. V primeru izbire desetih iteracij, dobimo za skoraj vse kombinacije pristopov in podatkovnih zbirk boljše rezultate v primerjavi z osnovnimi ocenami. Medtem, ko so rezultati pri petih in petnajstih iteracijah močno odvisni od izbrane zbirke. V primeru zbirke XQLFW je očitna boljša izbira manjšega števila iteracij, za zbirko CALFW pa večje število iteracij.

\section{Zaključek}\label{sec:conclusion}

V članku smo predstavili izviren postopek optimizacije predhodno izračunanih ocen kakovosti obraznih slik, ki skuša v začetne ocene kakovosti vnesti dodatno informacijo, pridobljeno iz distribucije podobnost ujemajočih se parov slik. Rezultati eksperimentov s tremi metodami za oceno kakovosti in tremi podatkovnimi zbirkami (CP\-LFW, CALFW in XQLFW) kažejo na močno odvisnost uspešnosti postopka od izbire števila iteracij optimizacije. V primeru manjšega števila iteracij, so rezultati za podatkovni zbirki CPLFW ter CALFW slabši od osnovnih ocen kakovosti. V primeru prevelikega števila iteracij se zgodba ponovi za zbirko XQLFW. Najboljše rezultate dosledno ustvari pristop pri desetih iteracijah, kjer so rezultati večinoma boljši od osnovnih ocen izbranih pristopov.

\small
\bibliographystyle{ieee}
\bibliography{erk}

\end{document}